\documentclass[10pt,twocolumn,letterpaper]{article}
\usepackage{amsmath}
\usepackage{amsfonts}
\usepackage{amssymb}
\usepackage{pifont}

\usepackage{dblfloatfix}
\usepackage{gensymb} 

\usepackage{graphicx}   %
\usepackage{wasysym}    %
\usepackage{booktabs}   %
\usepackage{xcolor}     %
\definecolor{hbblue}{HTML}{6BA8D6}
\definecolor{hbgreen}{HTML}{00A857}
\newcommand{\hbf}{\textcolor{hbgreen}{\CIRCLE}}      %
\newcommand{\hbh}{\textcolor{hbgreen}{\LEFTcircle}}  %
\newcommand{\hbe}{\textcolor{hbgreen}{\Circle}}      %

\newcommand{\rd}[1]{\rotatebox{70}{#1}}

\usepackage{caption}

\usepackage[final,datasets]{wacv}      %
\definecolor{wacvblue}{rgb}{0.21,0.49,0.74}
\usepackage[pagebackref,breaklinks,colorlinks,allcolors=wacvblue]{hyperref}

\title{\textsc{Allude}: A Unified Evaluation System for Configurable Attacks in Differentiable Environments}

\author{Mansi Phute$^1$ \and 
Alexander Greenhalgh$^1$ \and 
Matthew Hull$^1$ \and
Haoran Wang$^1$ \and 
Alec Helbling$^1$ \and 
ShengYun Peng$^1$ \and 
Elliott Faa$^1$ \and 
Willian Lunardi$^2$ \and 
Martin Andreoni$^2$ \and 
Wenke Lee$^1$ \and
Duen Horng Chau$^1$ 
}

\usepackage{graphicx}

\usepackage{booktabs}
\usepackage{url}
\usepackage{xcolor}
\usepackage{multirow}
\usepackage{nicefrac}
\usepackage{multirow}
\usepackage{multicol}
\usepackage{hyperref} 
\usepackage{pdflscape}
\usepackage{adjustbox}

\hypersetup{
    colorlinks,
    linkcolor={blue!80!black},
    citecolor={blue!80!black},
    urlcolor={blue!80!black}
}
\definecolor{lightyellow}{RGB}{255, 255, 190} 

\newcommand{\tool}{\textsc{Allude}\xspace}

\newcommand{\paraheader}[1]{\par\vspace{-\parskip}\vspace{4pt}\noindent\textbf{#1}}

\begin{document}

\twocolumn[{
\maketitle
\vspace*{-0.25in}
\centering

\begin{center}
$^{1}$ Georgia Institute of Technology, USA\\
$^{2}$ Technological Innovation Institute, UAE
\end{center}

\includegraphics[width=0.98\linewidth]{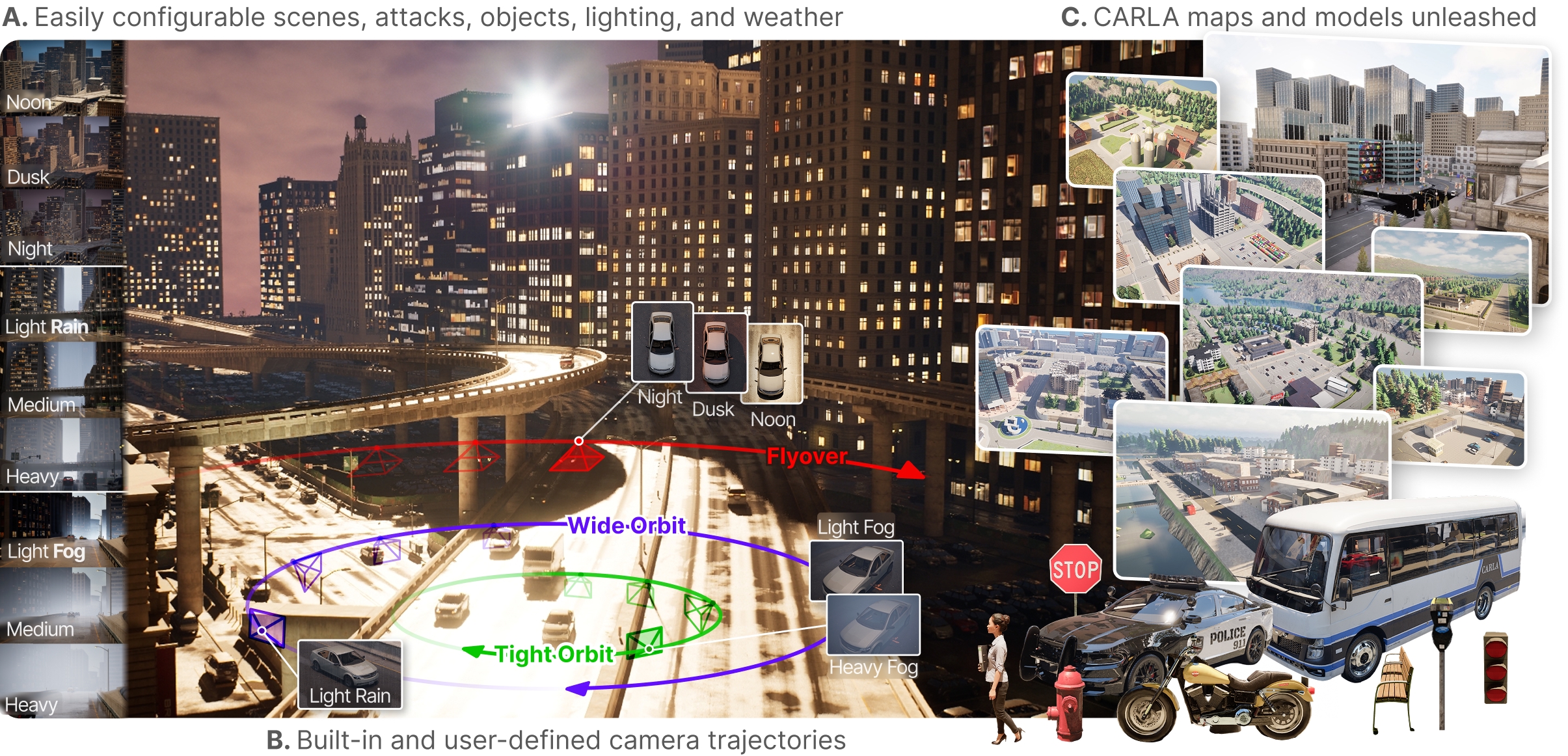}
\vspace{-0.5em}
\captionof{figure}{\tool{} closes long-standing gaps in adversarial attack evaluation, uniting Unreal Engine photorealism with end-to-end differentiable rendering in a single cross-platform system.  
\textbf{(A)} fully configurable scenes, objects, attacks, lighting, and weather;
\textbf{(B)} a rich set of built-in camera trajectories, from tight orbits to sweeping flyovers;
and \textbf{(C)} a modernized library of CARLA maps and models, freed from legacy Unreal Engine versions for immediate reuse in current and future research.}
\label{fig:crownjewel}
\vspace{0.7em}
}]

\maketitle

\begin{abstract}

Adversarial attacks against vision models like object detectors are often evaluated under limited conditions, leaving their performance under-characterized. Bridging simulation and differentiable rendering enables more robust, end-to-end evaluation of these adversarial attacks, yet there is no easy-to-use, unified system that offers a rich set of customizable configurations for adversarial attacks across multiple scenes, objects, environmental and lighting conditions, and camera trajectories. 
We present \tool, which addresses these gaps, offering first-of-its-kind evaluation capabilities across Linux and Windows.
We comprehensively demonstrate \tool{}'s evaluation breadth through a two-pronged strategy: (1) using Latin Hypercube Sampling, we draw a representative subset from 5,400 configurations spanning 10 scene–object pairs, 9 weather conditions, 4 optimizers, 5 camera trajectories, and 3 detection models;
(2) we stress-test existing attacks (CAMOU, RAUCA, FCA) under diverse weather conditions and continuous camera trajectories, revealing degradation of attack success across every attack, exposing evaluation gaps in prior work. Through \tool{}'s end-to-end differentiable rendering, adversarial attacks can be optimized against shifting real-world deployment conditions. Our cross-platform open source code is available at \url{https://anonymous.4open.science/r/ALLUDE-A3F6}.

\end{abstract}

\section{Introduction}

\begin{figure}[t]
    \centering
    \includegraphics[width=0.95\linewidth]{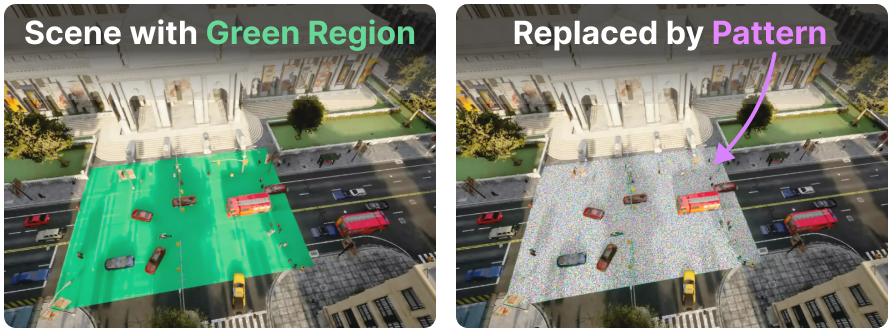}
    \caption{Existing adversarial attack evaluation ``superimposes'' the adversarial pattern onto the pre-determined placeholder region (here: green patch). Any overlapping objects (here: cars, shadows), are identified and replaced using their metadata. While this appears as true pattern replacement, the adversarial pattern is not rendered in the environment, thus preventing accurate evaluation.}
    \label{fig:patch}
\end{figure}

Robustness of vision models is an important area of research, providing the backbone to many safety-critical applications that depend on reliable and consistent visual predictions from modern detectors. 
Across adversarial attack literature, targeted perturbations in images on both pixel-space and 3D textures result in misclassification and detector-suppression attacks on a diverse set of objects, including vehicles \cite{hull2025complicitsplat, zhang2018camou, das_wang2024adversarialexamplesphysicalworld, FCA_2022, Suryanto_2023_ICCV, zhou2024rauca}, traffic signs \cite{Chen_2019}, and pedestrians \cite{wei2024physical, xu2020adversarialtshirtevadingperson}. 
Adversarial patterns, or ``patches'' are typically generated by iterative optimization, where noise is added to an image and iteratively optimized against a model's object-class predictions to induce classification or detection failure. 
Researchers often use photorealistic simulations to better understand how these patches perform in physical settings and in varying environments and weather conditions that would not be possible otherwise. 
Currently, patches are not inserted into the simulation but rather added onto the rendered scenes during postprocessing \cite{wei2024physical}. 
During scene rendering, the area adversarial patches will occupy is filled with a colored placeholder, like a ``green screen'' in \autoref{fig:patch} that is visually distinct from its environment. The corners of this plain patch are then calculated in postprocessing with the optimized adversarial patch superimposed on the image using PyTorch transformations \cite{wei2024physical}.

This postprocessing approach leads to two key issues: (1) The dynamic coordinate detection required for depositing patches onto the placeholder texture needs to be consistent across frames in the same scene; in practice, this is rarely the case. An error of a couple pixels can lead to the inserted adversarial patch appearance differing drastically across frames (\autoref{fig:patch}). Under the intended threat model, where an adversarial patch is deployed in the physical world, the patch's appearance should be consistent across views in the same scene. Frame-to-frame inconsistency violates this assumption, and causes evaluation approaches to overestimate attack strength and underestimate defense robustness. (2) The optimized pattern does not respond properly to environmental constraints, such as correct lighting or physics, which can cause the adversarial patch to fail when inserted into photorealistic simulation \cite{wei2024physical}. 

\begin{table*}[t!]
\centering
\caption{\tool{} (bottom row) is the only adversarial attack framework with full marks in every object domain and system capability, uniquely combining arbitrary geometry, engine-faithful rendering, configurable environments, cross-platform support, and open-source release.
The object-domain columns mark the classes each method is demonstrated on; 
``Other'' denotes classes beyond the three often-studied types (vehicle, pedestrian, traffic sign); \tool{} supports arbitrary simulation objects. 
\textbf{Capabilities}: \hbf{}~full, \hbh{}~partial, \hbe{}~none.}
\label{tab:capability_matrix}
\renewcommand{\arraystretch}{1.15}
\setlength{\tabcolsep}{4pt}
\footnotesize
\begin{tabular}{l l l l cccc ccccc}
\toprule
& & & & \multicolumn{4}{c}{\textbf{Object domains}} & \multicolumn{5}{c}{\textbf{Capabilities}} \\
\cmidrule(lr){5-8} \cmidrule(lr){9-13}
\textbf{Method} & \textbf{Param.} & \textbf{Renderer} & \textbf{Sim.} & \rd{Vehicle} & \rd{Pedestrian} & \rd{Traffic sign} & \rd{Other} & \shortstack{Arbitrary\\geometry} & \shortstack{Engine-\\faithful\\rendering} & \shortstack{Config.\\environ.} & \shortstack{Cross-\\platform} & \shortstack{Open\\source} \\
\midrule
CAMOU~\cite{zhang2018camou} & Tiled 2D & Clone network & UE4 & \hbf & \hbe & \hbe & \hbe & \hbh & \hbh & \hbh & \hbe & \hbf \\
DAS~\cite{das_wang2024adversarialexamplesphysicalworld} & Partial 2D & Neural renderer & CARLA (UE4) & \hbf & \hbe & \hbe & \hbe & \hbh & \hbh & \hbe & \hbe & \hbf \\
FCA~\cite{FCA_2022} & Full UV & Neural renderer & CARLA (UE4) & \hbf & \hbe & \hbe & \hbe & \hbh & \hbh & \hbh & \hbe & \hbf \\
ACTIVE~\cite{Suryanto_2023_ICCV} & Triplanar & Neural renderer & CARLA (UE4) & \hbf & \hbe & \hbe & \hbe & \hbf & \hbh & \hbh & \hbe & \hbe \\
RAUCA~\cite{zhou2024rauca} & Full UV & Neural renderer & CARLA (UE4) & \hbf & \hbe & \hbe & \hbe & \hbh & \hbf & \hbh & \hbe & \hbf \\
Adv.\ T-shirt~\cite{xu2020adversarialtshirtevadingperson} & 2D patch & --- & --- & \hbe & \hbf & \hbe & \hbe & \hbe & \hbe & \hbe & \hbe & \hbf \\
Invis.\ Cloak~\cite{wu2020invisibilitycloak} & 2D patch & --- & --- & \hbe & \hbf & \hbe & \hbe & \hbe & \hbe & \hbe & \hbe & \hbf \\
ShapeShifter~\cite{Chen_2019} & 2D pert. & --- & --- & \hbe & \hbe & \hbf & \hbe & \hbe & \hbe & \hbe & \hbe & \hbf \\
\midrule
\textbf{\tool{} (Ours)} & Full UV & \textbf{Mitsuba 3}~\cite{jakob2022mitsuba3} & \textbf{UE5} & \hbf & \hbf & \hbf & \hbf & \hbf & \hbf & \hbf & \hbf & \hbf \\
\bottomrule
\end{tabular}
\vspace*{-0.1in}
\end{table*}

As summarized in \autoref{tab:capability_matrix}, existing attacks often couple a specific texture parameterization, differentiable renderer, target detector, and simulator into a  fixed attack pipeline; 
attacks are typically demonstrated on a single object domain in isolation, rendered through standalone neural renderers, optimized against legacy detectors, and run only on the simulator's native platform. 
No existing system offers engine-faithful rendering, configurable weather and camera trajectories, or cross-platform support within one generalized framework. 
This fragmentation makes attacks difficult to compare or evaluate under diverse, deployment-relevant conditions that matter for safety-critical vision applications.

To close these long-standing gaps in adversarial attack evaluation, we introduce \tool (\autoref{fig:crownjewel}), a system that allows easy cross-platform development on Linux along with Windows systems, making adversarial research more accessible and easier to integrate with existing research pipelines (\autoref{tab:capability_matrix}, bottom row). \tool enables easy customization of environments and provides environment settings for various weather and lighting conditions. 
In addition, we modernize a broad set of CARLA \cite{dosovitskiy2017carla} assets by uncoupling them from their original UE version. 
These CARLA environments and assets are often used in adversarial attacks and data generation. 
However, they are tied to the UE version CARLA uses, preventing researchers from using these assets in newer UE versions. 
By separating CARLA objects from their UE versions, we allow greater customization and improve usability for the adversarial machine learning research community. 
The main contributions of \tool are:
\begin{enumerate}
    \item \textbf{\tool{}: Easy-to-use, and cross-platform system with a rich set of customizable configurations for adversarial attacks, scenes, objects, environmental and lighting conditions, and camera trajectories} (\autoref{fig:crownjewel}; \autoref{sec:system}). 
    \tool{} readily integrates into existing research and evaluation pipelines (demonstrated in \autoref{sec:lit_attacks}), offering the first-of-its-kind evaluation capabilities across Linux and Windows.

    \item \textbf{Two-pronged evaluation strategy for comprehensive testing of \tool{}:} We demonstrate \tool{}'s generalization through prototypical attack settings based on prior literature, sampling a representative subset of $N=100$ configurations through \textit{Latin Hypercube Sampling} from over 5,400 possible configurations, isolating the effects of detector model, environmental conditions and camera trajectories on attack performance.
    
    \item \textbf{Open source implementation:} Our code, source data, and generated data are all open source at {\small \url{https://anonymous.4open.science/r/ALLUDE-A3F6}}. We also make the CARLA assets uncoupled from CARLA's Unreal Engine (UE) version available at {\small \url{https://huggingface.co/datasets/allude-occluded/real-carla-assets-anon}}.
    \end{enumerate}

\section{Related Work}
\label{sec:related-work}

\paraheader{Simulation in Adversarial Machine Learning.}
Simulation is an important aspect of optimizing adversarial attacks to test systems \cite{wu2020physical, zhang2024visual, nesti2022evaluating}. 
Unreal Engine is the backbone of many simulators including CARLA \cite{dosovitskiy2017carla} and Airsim \cite{shah2017airsim}. It is selected for its high fidelity and photorealistic rendering. 
However, Unreal Engine is non-differentiable, making direct optimization impossible and preventing it from seamlessly fitting into the adversarial optimization pipeline. This also persists in all simulation benchmarks derived from them \cite{fonder2019midair, wang2020tartanair, lin2022capturing, rizzoli2023syndrone, khose2024skyscenes, meier2024carladrone}.
Adversarial approaches rely on approximations such as post-hoc superimposition of adversarial textures onto rendered images, which can leave errors (\autoref{fig:patch}). Inserting a texture in simulation breaks attacks \cite{xu2024imperceptible} and affects attack transferability \cite{nesti2022evaluating}. 

\noindent
\paraheader{Differentiable Rendering.}
Differentiable rendering allows us to merge traditional computer graphics and gradient-based optimization. It allows backpropagation of gradients to update geometry, lighting, and textures in a scene ~\cite{loper2014opendr, kato2017neural, laine2020modular}.
In adversarial machine learning differentiable rendering is used to optimize textures in varying environmental conditions ~\cite{athalye2018synthesizing, zhou2024rauca, li2024flexible, jia2025cca, hull2025complicitsplat}.
However, a limitation of these methods, such as Mitsuba and PyTorch3D, is that they do not support simulation or complex weather and require specialized knowledge to use. 

\noindent
\paraheader{Bridging Simulation and Differentiable Rendering.}
Adversarial attacks span multiple object domains, such as vehicles \cite{hull2025complicitsplat, zhang2018camou, das_wang2024adversarialexamplesphysicalworld, FCA_2022, Suryanto_2023_ICCV, zhou2024rauca}, traffic signs \cite{Chen_2019}, and pedestrians \cite{wei2024physical, xu2020adversarialtshirtevadingperson}. As shown in \autoref{tab:capability_matrix}, each combines texture parameterization, differentiable renderer, and target detector into a method-specific attack pipeline. Vehicle camouflage methods utilize differentiable rendering over different 2D texture representations, tying their evaluation to specific renderers and legacy detection models for consistency with existing literature. Pedestrian and sign adversarial attacks avoid 3D differentiable rendering entirely, operating on only 2D patches that suffer from the same issues shown in \autoref{fig:patch}. 
These discrepancies between individual attack pipelines limit cross-framework reproducibility, where a texture optimized through one method's renderer and UV mapping cannot be evaluated without manual reunification of their attack pipeline component. 
Recent work bridges differentiable rendering with photorealistic simulation, enabling end-to-end optimization of adversarial textures on arbitrary 3D objects within controllable environments~\cite{phute2025undream}. This establishes that in-simulation attacks are  \textit{feasible}, but stops at demonstrating the capability --- on Windows, a handful of objects and conditions, and a single detector --- leaving open the empirical question this capability makes answerable, when do physical attacks actually succeed, and when does success measured under narrow conditions fail to generalize? \tool{} turns this capability into the first systematic study of that question, contributing (i) the cross-platform infrastructure (headless Linux) needed to run it at scale, (ii) a standardized configuration space and quantitative characterization of attack behavior across it, and (iii) a released, version-decoupled CARLA asset library so the evaluation is reusable. 
\section{System}
\label{sec:system}
\tool supports a wide range of evaluation configurations, with presets for 10 objects, 10 scenes, 5 trajectories, 9 weather conditions, 4 optimizers, and 3 detectors (\autoref{sec:usability}).
\tool enables optimizing adversarial perturbations of arbitrarily shaped objects on Linux and Windows systems without requiring a graphical user interface (GUI) (\autoref{sec:cross-platform}). This brings UE's photorealistic rendering to headless Linux systems such as HPC clusters and AWS servers, enabling large-scale, distributed evaluations of adversarial attacks in realistic simulation environments. It also allows integration of CARLA scenes and assets in experiments 
while eliminating their dependency on legacy UE versions and the CARLA stack
(\autoref{sec:carla}).

\subsection{Easy to Use \& Customize}
\label{sec:usability}

\subsubsection{Diverse Evaluation Configurations}
\label{sec:customization}

\begin{figure}[t]
    \centering
    \includegraphics[width=\linewidth]{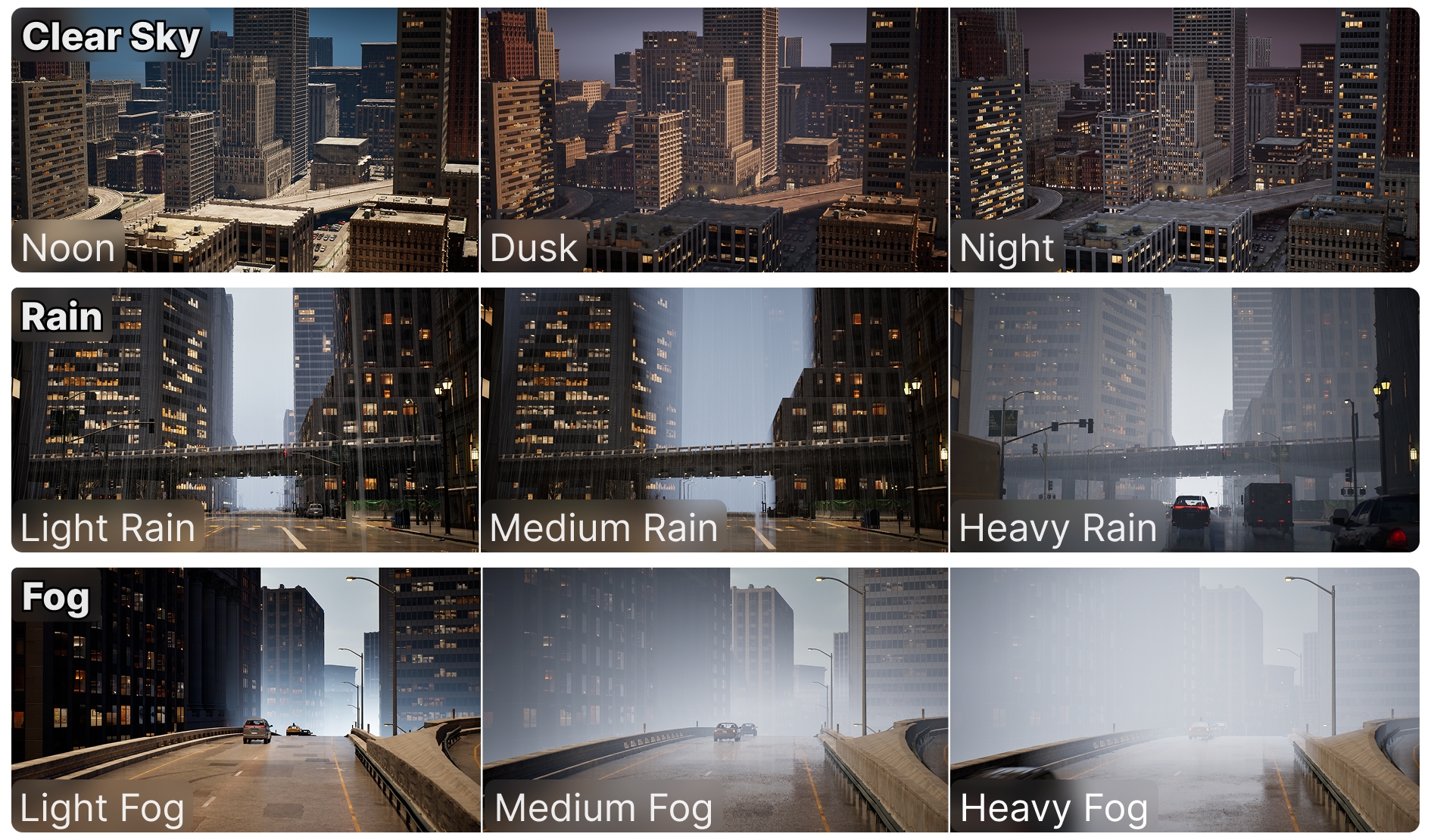}
    \caption{\tool offers easy weather and lighting customization for stress-testing attacks. Users can control lighting to reflect time of the day, and add rain and fog with varying intensity levels.}
    \label{fig:custom-weather}
\end{figure}

\paraheader{Weather:} A key advantage of using simulation in evaluating adversarial attacks is the wide variety of environments and weather conditions that can be replicated, allowing greater control over the experimental setup. \tool takes advantage of Unreal Engine's suite of plugins and tools to include experiments with a variety of lighting and weather conditions. \tool spans 9 weather conditions across three categories: clear skies at three times of day (\emph{noon}, \emph{dusk}, and \emph{night}), three intensity levels each of rain and fog (\emph{light}, \emph{medium}, and \emph{heavy}) (\autoref{fig:custom-weather}). Full Unreal Engine per-condition weather and lighting parameters are provided in \autoref{supp:weather}.

\begin{figure}[t]
    \centering
\includegraphics[width=\linewidth]{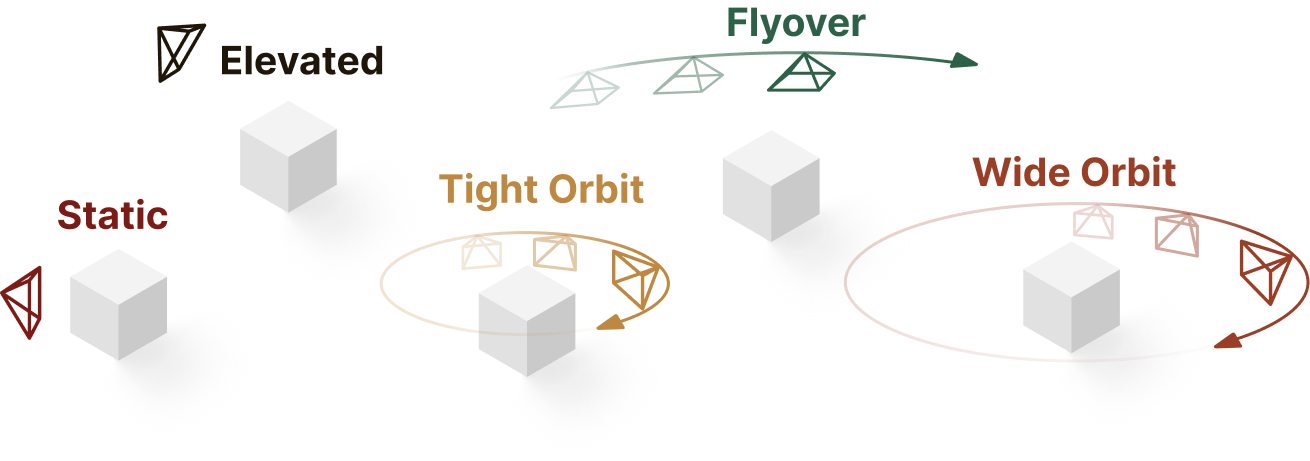}
    \caption{\tool{}'s 5 camera trajectories (static frontal,  static elevated, flyover, tight orbit, wide orbit) optimize and evaluate adversarial textures across viewpoints that existing approaches miss.}
    \label{fig:trajectory}
\end{figure}

\paraheader{Trajectories:} \tool provides 5 camera trajectories: \emph{static frontal}, \emph{static elevated}, \emph{flyover}, \emph{tight orbit}, and \emph{wide orbit}, designed to be representative of how cameras capture objects in real-world deployment settings. These trajectories are illustrated in \autoref{fig:trajectory}. Unlike prior adversarial simulation benchmarks~\cite{LIAN2025114395} that sample from static viewpoints, \tool{} supports creation of continuous camera trajectories. Detailed per-trajectory camera parameters and object-specific scale transformations are given in \autoref{supp:trajectories}.

\paraheader{Optimizers:} \tool{} implements four gradient-based adversarial optimizers: FGSM \cite{goodfellow2015explainingharnessingadversarialexamples}, PGD \cite{madry2018towards}, Auto-PGD, and AutoAttack \cite{croce2020reliable}. While originally designed for $\ell_\infty$-bounded attacks on image classifiers (2D pixel space), in \tool{}'s differentiable-rendering setting they serve as the optimizer component of the 3D texture-space attack. Per-optimizer implementation details and hyperparameters are included in \autoref{supp:optimizers}.

\paraheader{Detectors:} \tool{} provides built-in support for 3 popular detection models: YOLOv11, Faster R-CNN, and DETR, with the option to add in support for more. The selected models cover modern one-stage, two-stage detectors, and transformer-based architectures,
enabling the framework's detector capabilities to be explored across many evaluation settings. For example, 
the Latin hypercube analysis (\autoref{sec:lhs}) easily varies the detector as one of the configurations. 

\subsubsection{Ease of Use}
\tool provides a comprehensive suite of options for configuring the various customizations described in \autoref{sec:customization}. It runs as a single-line CLI command on both Linux and Windows, enabling quick adjustments to experiments. All parameters including attack hyperparameters, weather, and environment can be modified directly through the CLI.

\subsection{First-Of-Its-Kind Cross-Platform Evaluation}
\label{sec:cross-platform}
\tool{} provides researchers with cross-platform Linux and Windows support not previously available. 
Historically, Unreal Engine has been used primarily on Windows due to its more intuitive graphical user interface (GUI) and stronger user support.
While a Linux distribution for Unreal Engine is available, it is not widely used. 
For data generation and attack evaluation in adversarial research, Linux compatibility is necessary to integrate evaluation capabilities into researchers' current research pipelines, which heavily favor headless Linux-based GPU servers. 
The headless setting of these GPU servers introduces further complications in rendering photorealistic simulations:
Windows applies sRGB gamma encoding through the display swapchain automatically, while headless Linux saves linear-light pixels without that conversion. However, it is possible to force sRGB output encoding in Linux, and we provide a docker configuration in our code-base for reproducibility.

\subsubsection{Cross-platform Runtime Stability Engineering}
When a Python callback receives a borrowed UStruct argument and releases the Global Interpreter Lock (GIL) for an extended period, which occurs in \tool when a multi-second subprocess call is run, the garbage collector in Unreal Engine traverses the wrapper after its backing memory has been reclaimed. This results in a fatal Unreal Editor crash. We mitigate this with a three-part fix extending wrapper validity, pinning object references, and modifying garbage collection mechanisms.
The complete patch description is provided in the supplementary \autoref{supp:runtime-patch}.

\subsubsection{
Isolating independent C++ backbones}
Running a real-time game engine with a differentiable renderer and a deep-learning framework in a single Python process faces three independent toolchain conflicts on Linux due to C++ conflicts.
We therefore separate these components into two cooperating simultaneous processes. The engine drives scene capture in the parent process; a child process performs gradient computation and texture optimization. Additional details about this implementation can be found in the supplementary \autoref{supp:two-process-arch}.

\subsection{Decoupling CARLA assets from Legacy Unreal}
\label{sec:carla}
CARLA is a self-driving simulation platform built on Unreal Engine, widely used in data generation pipelines for adversarial attacks. It inherits many of UE's simulation capabilities but is very tightly coupled to a specific UE version, making it difficult to update with newer releases of Unreal Engine. Using or modifying CARLA assets in Unreal also requires installing CARLA from source, which is time consuming and tedious. In all experiments in \autoref{sec:lit_attacks} and \autoref{sec:lhs}, we use CARLA assets and environments upgraded to Unreal Engine 5.7 and uncoupled from CARLA backend dependencies. We open-source these modernized assets to enable researchers to customize CARLA assets easily. Additional details on modernizing these assets are provided in \autoref{supp:carla-assets}, and the assets are available at \url{https://huggingface.co/datasets/allude-occluded/real-carla-assets-anon}.
\section{Evaluation}

We evaluate \tool{}'s capabilities through two complementary approaches, 
(1) demonstrating the broad attack capabilities through sampling a subset of the 5,400 evaluation configurations through Latin Hypercube Sampling (LHS) to characterize how targeted attack object, optimizer, detector, trajectory, and weather conditions influence attack success across the full configuration space (\autoref{sec:lhs});
and
(2) stress-testing \tool{}'s optimized adversarial textures in the weather and trajectory conditions specified in \autoref{sec:customization} on the Audi E-Tron vehicle (\autoref{sec:lit_attacks}).

\paraheader{Metric: } Across both studies we report adversarial mAP@50 (adv-mAP@50) of the target class, where lower values indicate a stronger attack. 
Prior physical attack literature commonly reports attack success rate (ASR), calculated as the fraction of frames where a target is no longer detected. 
This metric carries the assumption of perfect benign detection, yet several of our weather and trajectory conditions degrade benign detection: benign mAP@50 falls to $0.89$ under the flyover trajectory (\autoref{tab:etron}), and $14$ of the $100$ sampled LHS configurations are \emph{indeterminate}, with the object undetected even without an attack (\autoref{sec:lhs}). 
To avoid conflating attack effect with environmental detection failure, we use mAP@50, which is the mean average precision of the target-class detections at an intersection-over-union (IoU) threshold of 0.5:

\begin{equation}
\text{adv-mAP@50} = \frac{1}{|C|}\sum_{c \in C} \text{AP}_c^{\text{IoU}=0.5},
\end{equation}

With diverse environmental conditions and a broad range of object classes, mAP@50 provides a single comparable measure of attack strength that remains meaningful even when benign object detection is imperfect.

\subsection{Evaluating Diverse Configurations via Latin Hypercube Sampling}
\label{sec:lhs}

\begin{figure}[t]
    \centering
    \includegraphics[width=\linewidth]{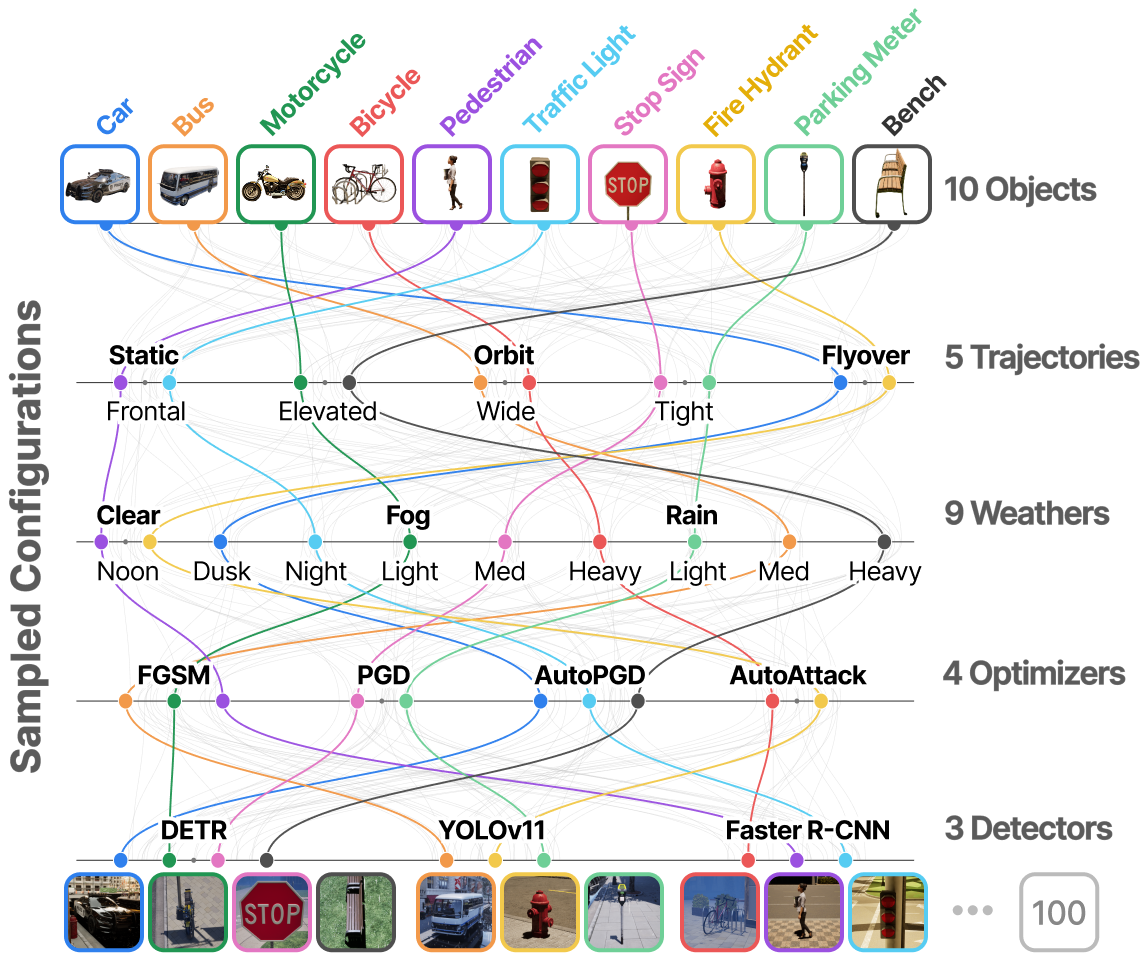}
    \caption{$N=100$ configurations through balanced Latin Hypercube Sampling (LHS) over 4 optimizers, 10 scene-object pairs, 5 trajectories, 9 weather conditions, and 3 detectors.}
    \vspace*{-0.2in}
    \label{fig:LHS_Configurations}
\end{figure}

To characterize attack behavior across the broader configuration space, we sample $N=100$ configurations through balanced Latin Hypercube Sampling (LHS) over 5 categorical factors: optimizer (4 types), scene-object pairs (10 levels), trajectory (5 levels), weather (9 levels), and detector architecture (3 levels) with a deterministic seed. 
The full factor space contains $5,400$ cells, with the $N=100$ sample covering $1.85\%$ of the space (\autoref{fig:LHS_Configurations}). The balanced marginals allow estimation of each factor's main effect on mAP@50; resolving interaction effects would require additional sampling fidelity within the configuration space. 
Every cell uses identical attack hyperparameters ($\epsilon=128$, $\alpha=\epsilon/8$, and attack-specific iteration counts). We perform an untargeted evasion attack. This experimental design samples a subset of the parameter space that is representative of the broader configurations enabled by \tool, demonstrating the framework's extensibility to a wide variety of adversarial attack conditions. Further details of the LHS approach, including renders of all $N=100$ scenes used in this analysis are provided in \autoref{supp:evaluation}.

\paraheader{Scenes:} Scenes are drawn from CARLA town and city environments from the Unreal Engine Fab store. Of CARLA's 10 towns, Towns 8 and 9 are absent from the open-source release and are excluded. 
Towns 1 and 2 are near-duplicates: a small village and a small town, where a single scene is representative. 
Town 3 is an urban layout similar to Town 10, except with reduced asset fidelity. 
We select one scene from each of Towns 2, 4, 5, 6, and 7, and five scenes from Town 10. This selection balances two goals: (1) scene diversity, as Towns 4 and 6 have less urban layouts than the other open-source towns, and (2) scene quality, as Town 10 offers the most realistic urban environments in CARLA. Example scenes from each town are shown in \autoref{supp:scene-details}.

\paraheader{Objects:} The 10 evaluation objects are sourced from CARLA and drawn from common COCO categories, matching the training distribution of the detection models: car, bench, traffic light, person, parking meter, bicycle, bus, fire hydrant, motorcycle, and stop sign. 
The literature comparison uses the Audi E-tron vehicle for consistent comparison with adversarial attack literature. Full scene and object details are described in \autoref{supp:object-details}. 

\paraheader{Attack outcome and model. } We define an attack as \emph{successful} when the adversarial texture suppresses the target-class detection score such that adv-mAP@50 $< 0.5$. 
Of the 100 cells, 14 are \emph{indeterminate}, where the detector fails to find the benign object (benign mAP@50 $= 0$) due to weather and trajectory specific detector degradation. 
We exclude these, leaving $N=86$ scored cells. Across these, \tool{} successfully attacks $40/86~(46.5\%)$ of the sampled configurations, and measurably reduces detection confidence ($\geq 0.10$ mAP@50 drop) in $72.1\%$ of cells. 
We fit an L2-regularized logistic regression of mAP@50 across these 5 factors, and measure each factor's contributions by the corresponding drop in model fit (McFadden pseudo-$R^2$ \cite{mcfadden1972conditional}) with its removal, with a likelihood-ratio test for significance (\autoref{tab:doe-factors}).

\begin{table}[t]
\centering
\caption{Per-object attack outcomes across the $N{=}86$ scored LHS cells.
\emph{Suppressed} cells have adv-mAP@50 $<0.5$;
\emph{Degraded} cells have a measurable confidence reduction
($\geq 0.10$ mAP@50 drop). Rates span the full range, from universally
attackable (traffic light) to universally robust (fire hydrant).}
\label{tab:object-success}
\renewcommand{\arraystretch}{1.15}
\setlength{\tabcolsep}{6pt}
\footnotesize
\begin{tabular}{@{}l c c c c@{}}
\toprule
& \multicolumn{2}{c}{\textbf{Suppressed} ($<0.5$)} & \multicolumn{2}{c}{\textbf{Degraded} ($\geq 0.10$)} \\
\cmidrule(lr){2-3}\cmidrule(lr){4-5}
\textbf{Object} & \textbf{Cells} & \textbf{Rate} & \textbf{Cells} & \textbf{Rate} \\
\midrule
Traffic light  & 8 / 8  & 100\% & 8 / 8   & 100\% \\
Bus            & 8 / 10 & 80\%  & 10 / 10 & 100\% \\
Bicycle        & 5 / 8  & 63\%  & 7 / 8   & 88\%  \\
Motorcycle     & 5 / 8  & 63\%  & 8 / 8   & 100\% \\
Bench          & 2 / 4  & 50\%  & 4 / 4   & 100\% \\
Car            & 4 / 8  & 50\%  & 8 / 8   & 100\% \\
Parking meter  & 4 / 10 & 40\%  & 6 / 10  & 60\%  \\
Stop sign      & 3 / 10 & 30\%  & 8 / 10  & 80\%  \\
Pedestrian     & 1 / 10 & 10\%  & 2 / 10  & 20\%  \\
Fire hydrant   & 0 / 10 & 0\%   & 1 / 10  & 10\%  \\
\midrule
\textbf{Overall} & \textbf{40 / 86} & \textbf{46.5\%} & \textbf{62 / 86} & \textbf{72.1\%} \\
\bottomrule
\end{tabular}
\end{table}

\begin{table}[b]
\centering
\caption{Configuration factor contribution to adversarial attack success
($\text{adv-mAP@50}<0.5$) for \tool, across $N{=}86$ sampled configurations.
``Variance explained'' is the drop in model fit (McFadden pseudo-$R^2$) when the
factor is removed; ``Significant'' denotes a likelihood-ratio test at $p<0.05$
(object $p{=}0.002$, trajectory $p{=}0.020$, all others $p>0.50$).}
\label{tab:doe-factors}
\renewcommand{\arraystretch}{1.2}
\footnotesize
\begin{tabular}{@{}l c c@{}}%
\toprule
\textbf{Factor} & \textbf{Variance explained} & \textbf{Significant?} \\
\midrule
\textbf{Target object}   & \textbf{22\%}  & \textbf{Yes} \\
Camera trajectory  & 10\%     & Yes \\
Weather                  & 4\%            & No \\
Attack algorithm         & 2\%            & No \\
Detector                 & $<$1\%         & No \\
\bottomrule
\end{tabular}
\end{table}

\paraheader{Object class dominates attack success. } Target object class is the primary predictor of attack effectiveness, explaining $22\%$ of regression model fit ($p=0.002$). 
Attack success by object spans the full range (\autoref{tab:object-success}): traffic lights are suppressed in every non-indeterminate attack ($8/8$) and buses in most ($8/10$), whereas pedestrian ($1/10$) and fire hydrant classes ($0/10$) are the most difficult to attack. 
This variation arises from two distinct sources. 
(1) Prominent, texture dominated objects such as buses, traffic lights, and stop signs provide extensive adversarial surface and weaker shape priors \cite{geirhos2018imagenettrained}, while pedestrians and fire hydrants are detected largely from silhouette and contours that are inherently more difficult for a surface texture to disrupt \cite{xu2020adversarialtshirtevadingperson}. This robustness is especially pronounced for pedestrians, where detection models are trained on a wide distribution of poses.
(2) Objects differ in how their 2D UV map textures are rendered onto the 3D objects in Unreal.
Because of these differences, the effective perturbation budget that reaches the detector varies by object even at a fixed $\epsilon$. 
We characterize these per-object texture-handling differences in \autoref{supp:object-details}.

\begin{figure*}[t]
    \centering
\includegraphics[width=0.9\linewidth]{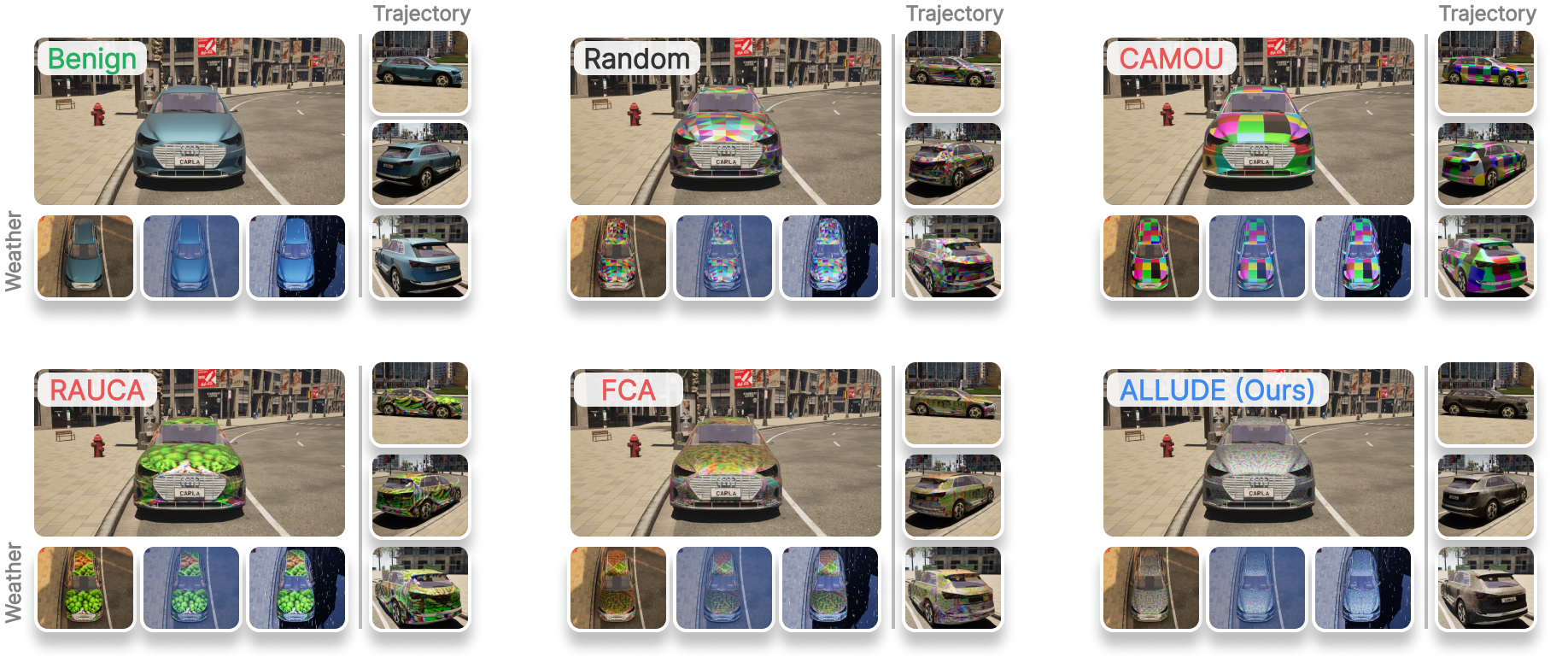}
    \caption{Comparison of adversarial textures on the Audi E-Tron under various environmental conditions. Baselines (Benign, Random), published attacks (CAMOU, RAUCA, FCA), and the \tool{}-optimized texture is rendered across weather conditions and camera trajectories. Textures that suppress detection at clear-noon static views (top) degrade as weather (right) and viewpoint (bottom) shift. 
    }
    \label{fig:literature_attacks}
    \vspace{-0.1em}
\end{figure*} 

\begin{table*}[t]
\centering
\caption{adv-mAP@50 of the target vehicle, \textbf{DETR}, on the Audi E-Tron literature comparison (lower $=$ stronger attack). Weather varied at static-elevated; trajectories at clear-noon. \tool{} is optimized natively against DETR (suppression objective, $\epsilon{=}255$) and applied unchanged across all conditions; the published textures are evaluated as transfer. Ground truth is the per-frame benign detection.}
\label{tab:etron}
\renewcommand{\arraystretch}{1.1}
\setlength{\tabcolsep}{4pt}
\footnotesize
\begin{tabular}{@{}l ccccccccc c cccc@{}}
\toprule
& \multicolumn{9}{c}{\textbf{Weather} (static-elevated)} & & \multicolumn{4}{c}{\textbf{Trajectory} (clear-noon)} \\
\cmidrule(lr){2-10} \cmidrule(lr){12-15}
\textbf{Condition} & \rotatebox{60}{Clear noon\,} & \rotatebox{60}{Dusk\,} & \rotatebox{60}{Night\,} & \rotatebox{60}{Rain L\,} & \rotatebox{60}{Rain M\,} & \rotatebox{60}{Rain H\,} & \rotatebox{60}{Fog L\,} & \rotatebox{60}{Fog M\,} & \rotatebox{60}{Fog H\,} & & \rotatebox{60}{Static front\,} & \rotatebox{60}{Flyover\,} & \rotatebox{60}{Tight orbit\,} & \rotatebox{60}{Wide orbit\,} \\
\midrule
Benign & 1.00 & 1.00 & 1.00 & 1.00 & 1.00 & 1.00 & 1.00 & 1.00 & 1.00 & & 1.00 & 0.89 & 0.94 & 1.00 \\
Random & 1.00 & 0.10 & 0.00 & 0.53 & 0.10 & 0.03 & 1.00 & 1.00 & 0.00 & & 0.96 & 0.64 & 0.70 & 0.95 \\
\midrule
CAMOU & 1.00 & 0.00 & 0.00 & 0.43 & 0.02 & 0.00 & 1.00 & 1.00 & 1.00 & & 1.00 & 0.76 & 0.70 & 0.97 \\
RAUCA & 0.00 & 0.00 & 0.00 & 0.00 & 0.00 & 0.00 & 0.00 & 0.00 & 0.00 & & 0.20 & 0.55 & 0.59 & 0.49 \\
FCA & 1.00 & 0.00 & 0.00 & 0.98 & 0.12 & 0.00 & 0.06 & 0.00 & 0.00 & & 0.53 & 0.82 & 0.76 & 0.86 \\
\midrule
\tool{} & 0.00 & 0.00 & 0.00 & 0.00 & 0.00 & 0.00 & 0.00 & 0.00 & 0.00 & & 0.33 & 0.78 & 0.76 & 0.95 \\
\bottomrule
\end{tabular}
\end{table*}
 
\paraheader{Viewpoint diversity drives attack robustness. } Camera trajectory is the second-strongest factor, explaining $10\%$ of model fit ($p=0.020$). Attack success across the sampled configurations decreases with trajectory complexity, where larger distributions of viewing angles and camera distances require the adversarial texture to remain effective across a wider range of conditions. Static trajectories, 
are the easiest to attack (static-elevated $78\%$, static-frontal $41\%$). Orbit trajectories circle the object through 360$\degree$, and their difficulty is consistent with viewing distance: the tight-orbit attacks succeed in $65\%$ of cells while the more distant wide-orbit succeeds in $45\%$. The detailed flyover trajectory combines wide angular coverage with varying distances and framing of the object, and is the hardest at $20\%$. Because this dependence on camera motion is only observable under continuous trajectories, static-viewpoint benchmarks \cite{LIAN2025114395, khose2024skyscenes} cannot expose it, yet it is among the strongest determinants of attack effectiveness.

\paraheader{Optimized textures remain effective across weather.} In the configuration space analysis, Weather explains $4\%$ of model fit and does not reach significance ($p=0.84$). 
This demonstrates that optimizing an effective adversarial texture is achievable across all weather conditions as long as benign detection remains reliable. 
Adversarial textures optimized in the various weather conditions in \tool{} reach the same rate of attack success across all nine weather conditions. 
Because each attack is optimized and scored under the same weather conditions, the optimizer always operates within its original training distribution. 
Attack success is shaped more by the target object and viewpoint than by atmospheric conditions, though we do not isolate weather directly in our experimental design.

\paraheader{Attack success generalizes across optimizers and detectors. } The attack algorithm and target detector are the two weakest factors in the design, explaining roughly $2\%$ and under $1\%$ of model fit and showing no significant differences across levels. The four optimizers (PGD, Auto PGD, AutoAttack and FGSM) perform at statistically indistinguishable rates ($41-57\%$; $p=0.51$), with overlapping confidence intervals at this sample size. 
The three detectors also show similar attack success rates across Faster R-CNN ($56\%$), DETR ($48\%$), and YOLOv11 ($38\%$), spanning one-stage, two-stage, and transformer architectures ($p=0.75$). 
The numerical ordering loosely tracks model recency: the oldest detector (Faster R-CNN) is most affected and the most modern (YOLOv11) least, but this trend is not statistically significant and we draw no conclusion from it; characterizing whether these modern architectures are systematically more robust would require a dedicated experimental design. 
Prior physical attack work has largely developed custom attacks tuned to the idiosyncrasies of individual object classes and detectors. 
\tool{} instead enables systematic evaluation across a broad range of objects, viewpoints, and conditions within one framework, which reveals where attack success generalizes and where it does not. 
For safety critical applications in autonomous driving, this distinction matters: robustness cannot be assessed against single objects, static viewpoints, or detector architectures in isolation, and conditions most relevant to deployment (continuous trajectories, degraded weather) are precisely those that prior static, single-object evaluations omit. 
By making such evaluation accessible across platforms, \tool{} provides the ability to characterize these critical safety vulnerabilities in simulation before they affect model deployment.

\subsection{Evaluating Published Attacks in ALLUDE} %
\label{sec:lit_attacks}
Published physical camouflage attacks couple their parameterization, rendering bridge, and loss into method-specific pipelines that do not transfer across frameworks.
We therefore evaluate the published adversarial textures themselves across 13 configurations (9 weather conditions with fixed frontal trajectory, 5 trajectories at clear-noon, one shared cell). 
These attacks are selected based on compatibility with \tool{}, as characterized by the per-method parameterization, renderer, and simulator in \autoref{tab:capability_matrix}. 
Faithfully transferring the UV maps of published textures into our pipeline requires re-unifying the vehicle's UV layout; details on this approach are provided in \autoref{supp:uv-reunification}.
We evaluate six textures: (1) the \textbf{Benign}, \textbf{Random} baselines; (2) adversarial textures \textbf{CAMOU}, \textbf{RAUCA}, and \textbf{FCA}; and (3) our \textbf{\tool{}} texture, optimized against DETR with a targeted suppression objective at $\epsilon{=}255$ on the Audi E-Tron vehicle (\autoref{fig:literature_attacks}). 
We optimize against DETR as a representative of \tool{}'s native optimization, not as an exhaustive or privileged choice. \tool{} is detector-agnostic; we select DETR because it is widely used in the detection literature and readily available through a standardized HuggingFace interface, simplifying integration and reproduction.
Because a manufactured physical texture cannot be re-optimized after deployment, we report each of the 13 conditions individually to characterize how each texture performs under environmental and viewpoint shift rather than inflating performance through condition-specific optimization. 

\paraheader{\tool{} achieves weather-invariant suppression.} 
Across all nine weather conditions, the \tool{}-optimized texture always suppresses mAP@50 to $0.00$, achieving complete evasion with a single fixed texture. 
Among the published attacks, RAUCA matches this weather-invariant suppression, while FCA and CAMOU transfer far less reliably: FCA fails at clear-noon ($1.00$) and under light rain ($0.98$), and CAMOU fails at clear-noon and across all three fog conditions, leaving it the weakest of the published literature attacks. 
This ordering, RAUCA strongest and CAMOU weakest, is consistent with the results reported by \cite{zhou2024rauca}, indicating that our rendering evaluation preserves the methods' established relative strengths. 

\paraheader{Continuous trajectories remain challenging for all attacks.} 
Under the moving camera trajectories (\autoref{tab:etron}), every texture weakens substantially, including RAUCA, whose trajectory-mean rises to $0.46$ on DETR from $0.00$ under variations in weather. 
No attack, published or ours, achieves evasion across the moving-camera views. 
Notably, FCA was designed to provide strong multi-performance \cite{FCA_2022}, yet rendered against DETR it does not retain that robustness, emphasizing the immense evaluation flexibility that \tool{} offers in making it possible to test against any number of user-defined trajectories to reveal blind-spots.
\tool{} reaches a trajectory-mean result of $0.71$ on DETR, with the largest reduction in the static-front view ($0.33$) while the flyover ($0.78$), tight-orbit ($0.76$), and wide-orbit ($0.95$) views remain largely detectable. 
This aligns with the \tool{} configuration space analysis of \autoref{sec:lhs}, where camera trajectory is the second-strongest factor in determining attack success, and demonstrates that viewpoint robustness measured from static viewpoints can overstate the effectiveness of existing adversarial attacks.

\section{Conclusion}
\tool{} is a first-of-its-kind system that is easy-to-use and offers a rich set of customizable configurations for adversarial attacks across multiple scenes, objects, environmental and lighting conditions, and camera trajectories, allowing cross-platform optimization on both Windows and Linux systems.
We comprehensively demonstrate \tool{}'s evaluation breadth through a two-pronged strategy.
We also make available CARLA assets such as objects and towns that can be imported and edited in the latest UE framework.
 {
    \small
    \bibliographystyle{ieeenat_fullname}
    \bibliography{main}
}
\clearpage
\setcounter{section}{0}
\renewcommand{\thesection}{S\arabic{section}}
\renewcommand{\thesubsection}{S\arabic{section}.\arabic{subsection}}
\renewcommand{\thetable}{S\arabic{table}}
\renewcommand{\thefigure}{S\arabic{figure}}
\setcounter{table}{0}
\setcounter{figure}{0}
\section*{Supplementary Material}

\begin{table*}[h]
\centering
\small
\caption{Shared Hyper parameters for all optimizers, identical across all detectors and 100 LHS cells used in evaluation.}
\label{tab:shared-hparams}
\begin{tabular}{llp{0.40\linewidth}}
\toprule
Parameter & Value & Additional Information \\
\midrule
$\epsilon$ ($L_\infty$ budget) & $128/255 \approx 0.502$ & pixel space $[0,1]$ \\
$\alpha$ (step size) & $\epsilon/8 = 16/255 \approx 0.0627$ & PGD / AutoPGD init \\
batch\_size & 10 & frames per gradient update \\
max\_batches & 4 & 40 frame-samples per outer iter \\
Frame sampling & stride-sampled & indices $[0, n/40, \dots, n-1]$ \\
clip\_min / clip\_max & 0.0 / 1.0 & post-step pixel range \\
$\epsilon$-ball projection & per-iter & $L_\infty$ around clean texture \\
AMP & enabled & \texttt{torch.cuda.amp.autocast} \\
Starting texture & random noise & uniform $[0,1]$ \\
Targeted? & False / True & cls (DETR, YOLO) / det (FRCNN) \\
\bottomrule
\end{tabular}
\end{table*}

\section{System Implementation Details}
\label{supp:system}

\subsection{Weather and Lighting Configurations}
\label{supp:weather}
\tool enables systematic evaluation of the attacks across nine distinct weather and lighting conditions, defined along three dimensions: (i) time of day (noon, dusk, and night), (ii) weather (rain, fog, and clear skies), and (iii) intensity (high, medium, and low).

\paragraph{Time of Day.}
The time of day is controlled primarily through the \texttt{DirectionalLight} object in Unreal Engine. By varying the light's angle, color, and intensity, we are able to reproduce the lighting characteristics corresponding to different times of day.

\paragraph{Weather:}
Weather effects are realized through a combination of Unreal Engine assets. (i) \textbf{Rain:} Rain is simulated using an asset constructed with Unreal Engine's Niagara particle system. Each such asset defines an emitter that spawns droplets within a volume above the scene, rendered using a combination of sprite and ribbon renderers, where the ribbon produces an elongated streak that replicates the effect of falling rain. Parameters such as the sprite spawn rate and velocity can be adjusted to control the appearance. (ii) \textbf{Fog:} Fog is simulated using Unreal Engine's \texttt{ExponentialHeightFog} asset, which exposes a range of parameters for controlling the fog's appearance. These include the primary and secondary fog layers, fog opacity, inscattering color, fog density, fog start and end distances, and fog falloff height.

\paragraph{Intensity:}
Adjusting the intensity of a given weather condition is not accomplished by varying a single parameter, but rather through the coordinated adjustment of several parameters. (i) \textbf{Rain:} The intensity of rain is increased by raising the spawn rate and velocity of droplets in the rain asset, increasing the fog density in \texttt{ExponentialHeightFog}, darkening the inscattering color slightly, and slightly reducing the light intensity of the \texttt{DirectionalLight} object. (ii) \textbf{Fog:} The intensity of fog is increased by raising the fog density, decreasing the fog height falloff, and increasing the fog opacity.

\subsection{Optimizer Configurations}
\label{supp:optimizers}

\tool{} integrates four optimizers into its adversarial attack pipeline: FGSM \cite{goodfellow2015explainingharnessingadversarialexamples}, PGD \cite{madry2018towards}, Auto-PGD, and AutoAttack \cite{croce2020reliable}. 
All four optimizers included with the initial release are $\ell_\infty$-bounded attacks that operate on the 2D UV texture of the target object. 
Every iteration, \texttt{torch\_worker.py} (1) renders a batch of frames through Mitsuba with the current texture as the parameter key, (2) computes the detector's loss on the rendered RGB, (3) backpropagates through the differentiable render to get the per-texel gradients on the UV texture, and (4) applies the update rule of the optimizer, then projects the perturbed texture back onto the $\ell_\infty$ $\epsilon$-ball around the baseline texture. Details on the shared optimizer hyperparameters are provided in \autoref{tab:shared-hparams}, while optimizer specific hyperparameters are shown in \autoref{tab:per-attack}.

\begin{table}[t]
\centering
\small
\caption{Per-attack algorithm details. Total calls per cell is the product of \texttt{max\_iter} with \texttt{max\_batches}.}
\label{tab:per-attack}
\begin{tabular}{lccp{0.36\linewidth}}
\toprule
Attack & max\_iter & Total calls & Update rule \\
\midrule
FGSM       & 1  & 4  & $x \gets x + \epsilon \cdot \mathrm{sign}(\nabla \mathcal{L})$ (single step) \\
PGD        & 10 & 40 & $x_{t+1} \gets \Pi_{\epsilon}\bigl(x_t + \alpha \cdot \mathrm{sign}(\nabla \mathcal{L})\bigr)$ \\
AutoPGD    & 10 & 40 & momentum-EMA on $\nabla \mathcal{L} / \mathrm{mean}(|\nabla \mathcal{L}|)$, step $\alpha \cdot \mathrm{sign}(\text{mom})$ \\
AutoAttack & 20 & 80 & 4-phase ensemble (Sec.~S1.2; APGD-CE $\to$ APGD-DLR $\to$ FAB $\to$ Square) \\
\bottomrule
\end{tabular}
\end{table}

\autoref{tab:autoattack-stages} provides additional detail on the 4-stage ensemble used in our AutoAttack implementation. A tracker of the best iteration (\texttt{\_aa\_best\_loss}) saves the highest-loss texture  across all  previous stages. When transitioning to the next stage in the ensemble, the algorithm restores this texture instead of utilizing a $2 \times \epsilon$ to prevent overstepping in the loss landscape. The final texture is the best loss texture across all four stages \cite{croce2020reliable}. The loss that drives each optimizer is detector-specific; \autoref{tab:detector-losses} summarizes the per-detector objective and task mode. DETR and YOLO (task=classification) are untargeted, while FRCNN (task=object detection) is targeted against the benign bounding box.

\begin{table}[t]
\centering
\small
\caption{AutoAttack 4-stage ensemble. Best-iterate snapshot restored on transition to next stage; the final returned texture is the best-loss iterate across all four phases \cite{croce2020reliable}.}
\label{tab:autoattack-stages}
\begin{tabular}{cclcp{0.20\linewidth}}
\toprule
Stage & Iters & Algorithm & Initial step & Step rule \\
\midrule
0 & 1--5   & APGD-CE  & $2\epsilon$    & halved when no improvement \\
1 & 6--10  & APGD-DLR & restored best  & same halving \\
2 & 11--15 & FAB      & $\alpha$       & boundary pursuit \\
3 & 16--20 & Square   & $\alpha$       & query-efficient random search \\
\bottomrule
\end{tabular}
\end{table}

\begin{table*}[b]
\centering
\small
\caption{Per-detector loss formulations used by the attack optimizer. DETR and YOLO operate under \texttt{task=cls} (untargeted); FRCNN under \texttt{task=det} (targeted suppression).}
\label{tab:detector-losses}
\begin{tabular}{llp{0.55\linewidth}}
\toprule
Detector & Task & Loss formulation \\
\midrule
DETR (\texttt{facebook/detr-resnet-50})
  & cls
  & Argmax-query cross-entropy: select query $j^* = \arg\max_q P(y | q_j)$, then $\mathcal{L} = \mathrm{CE}(\text{logits}[:, j^*, :], y)$. \\

YOLOv11n (ultralytics)
  & cls
  & Argmax-anchor sigmoid BCE: best-anchor selection, one-hot binary cross-entropy against target; sigmoid-probs cast to fp32 pre-BCE for AMP compatibility. \\

Faster R-CNN (\texttt{fasterrcnn\_resnet50})
  & det
  & Sum of all task losses in \texttt{model.train()} mode: $\mathcal{L} = \mathcal{L}_{\text{cls}} + \mathcal{L}_{\text{box}} + \mathcal{L}_{\text{obj}} + \mathcal{L}_{\text{rpn\_box}}$, with per-frame $\{\text{label}, \text{boxes}\}$ from \texttt{target\_det.json}. \\
\bottomrule
\end{tabular}
\end{table*}

\paraheader{Attack-region masks. } \tool{} can confine the optimized region to a binary UV mask, applying gradients only within the masked texels while the rest of the texture stays at its baseline value. Two objects use this (\autoref{tab:lhs-objects}): the pedestrian, masked to clothing via \texttt{rp\_mei\_clothes\_mask}, and the stop sign, masked to the sign face via \texttt{stop\_sign\_face\_mask}. These masks are optional, and all other objects are perturbed over their full UV texture.

\subsection{Camera Trajectory Parameters}
\label{supp:trajectories}

\tool{} ships five camera trajectories that govern how the object is viewed during optimization and evaluation. \autoref{tab:trajectories} lists each trajectory with its fixed frame count and camera-path description. Coordinates are in Unreal units (UU; 1\,UU $\approx$ 1\,cm), rotations in degrees (UE Euler), and camera FOV defaults to UE's 90$^\circ$.

\begin{table}[t]
\centering
\small
\caption{The five camera trajectories in \tool{}.  Frame counts are fixed by the \texttt{LevelSequence} asset per trajectory archetype, while the camera path is encoded as a per-frame transform list in \texttt{camera.json}.}
\label{tab:trajectories}
\begin{tabular}{lcp{0.50\linewidth}}
\toprule
Trajectory & Frames & Camera path \\
\midrule
static\_frontal    &  48 & Stationary; frontal viewpoint at object-centroid height. \\
static\_elevated   &  48 & Stationary; raised altitude with downward pitch. \\
tight\_attack\_orbit &  96 & Elliptical orbit, $\sim$5--10\,m radius around object, full 360$^\circ$. \\
wide\_orbit        & 120 & Elliptical orbit, $\sim$20--30\,m radius, full 360$^\circ$. \\
detail\_flythrough & 120 & Linear flyover, $\sim$6--30\,m traversal, varying pitch and distance. \\
\bottomrule
\end{tabular}
\end{table}

\subsection{Runtime Stability Patch}
\label{supp:runtime-patch}
While running \tool, we see issues in \texttt{PythonScriptPlugin} due to Python callback receiving a borrowed struct \texttt{FMoviePipelineOutputData} and the GIL is released for an extended period (e.g., during \texttt{subprocess.run()}). The Garbage Collector (GC) worker thread traces a now-dangling \texttt{StructInstance} and reads garbage \texttt{UObject} values, leading to a crash in \texttt{HandleObjectReference}. 
We mitigate the error by a two-pronged runtime patch: patching the plugin binary and managing garbage collection in the attack script. 
The only modification made on the source-level between the plugin  Python used in \tool and the default UE5 plugin is that we add a pointer validity check. We check if \texttt{StructInstance} is a dangling pointer holding a small garbage value, and if so the guard skips tracing instead of crashing in \texttt{HandleObjectReference}. 
The advantage of these runtime patches is that they do not require rebuilding Unreal Engine from scratch.

\subsection{Two-Process Architecture}
\label{supp:two-process-arch}
Running a Unreal Engine alongside Mitsuba and Pytorch in a single Python process incurs three independent toolchain conflicts on Linux: 
$(i)$ Divergent \texttt{libstdc$++$} revisions in the Unreal Engine binaries \texttt{libUnrealEditor-InputCore.so} and PyTorch prebuilt wheels are \texttt{libtorch\_python.so} loaded together causing duplicate C$++$ exception-handling symbols
$(ii)$ Incompatible CUDA runtime (PyTorch/Dr.Jit GPU stack) and threading runtime states (UE's Vulkan based renderer); and 
$(iii)$ Static OpenMP/TBB linkage in each component, which deadlocks when co-resident. 
Therefore, we separate the system into two cooperating processes,involving two independent Python interpreters and dependency stacks. UE interpreter drives scene capture and texture re-import in the parent process, while a separate conda interpreter is used for PyTorch and Mitsuba processes, performing gradient computation and texture optimization in a child process. The embedded interpreter never imports the GPU stack. Each iteration spawns a fresh child to which the attack configuration is passed as a serialized JSON message over \texttt{stdin}; and the child returns optimizer state which the parent imports. Because a new child is created and torn down on every attack iteration, its CUDA context is established and released cleanly each time, circumventing previously seen issues.

\subsection{Command-Line Interface Arguments}
\label{supp:cli}
To enable ease in experimentation, we allow the user to modify the parameters given to \tool from the CLI which is used to launch the attack. Detailed implementation instructiuon can be found in the github repository. The parts of the attack that can be changed by the CLI are: number of iterations, epsilon value, alpha value, source directory (which contains parameters such as original texture, scene information, XML configuration, and target object), along with Unreal Engine map information and the scene name in Unreal Engine. All other parameters such as detection model, detector task, number of frames in a sequence, attack type, Mitsuba to UE scale, maximum batches to allow optimization over (in case you have a batch limit), detection threshold, etc are exposed in \texttt{config/config.yaml}

\subsection{Modernized CARLA Assets}
\label{supp:carla-assets}

ALLUDE ships a UE 5.7–ported CARLA asset library, decoupled from the CARLA backend. Total: 18 GB on disk, 14,142 Unreal asset files (\texttt{13,912 *.uasset + 230 *.umap}). Additional information on each of these scenes is provided in \autoref{tab:carla-towns}, while additional information on the CARLA assets is given in \autoref{tab:asset-categories}.

\begin{table}[t]
\centering
\small
\caption{Modernized CARLA towns shipped with ALLUDE. Town01 and Town03 are included but unused in the LHS ablation (covered by Town02 and Town10HD respectively).}
\label{tab:carla-towns}
\begin{tabular}{llp{0.35\linewidth}}
\toprule
Town & File & Setting \\
\midrule
Town01    & Town01\_Opt.umap   & Small village \\
Town02    & Town02\_Opt.umap   & Small town \\
Town03    & Town03\_Opt.umap   & Urban \\
Town04    & Town04\_Opt.umap   & Rural \\
Town05    & Town05\_Opt.umap   & Suburban \\
Town06    & Town06\_Opt.umap   & Urban grid \\
Town07    & Town07\_Opt.umap   & Small town \\
Town10HD  & Town10HD\_Opt.umap & Dense urban \\
\bottomrule
\end{tabular}
\end{table}

Town08 and Town09 are absent from CARLA's open-source release.

\begin{table*}[h]
\centering
\small
\caption{Asset categories in the modernized CARLA project with 14,142 Unreal asset files including 13,912 \texttt{.uasset} and 230 \texttt{.umap}, $\sim$18\,GB on disk.}
\label{tab:asset-categories}
\begin{tabular}{lp{0.35\linewidth}r}
\toprule
Category & Path & Count / Size \\
\midrule
Maps                 & Content/Carla/Maps/             & 8 \texttt{.umap} \\
Object meshes        & Content/CV\_Pipeline/Meshes/     & 10 per-class \texttt{SM\_*.uasset} \\
Materials            & Content/CV\_Pipeline/Materials/  & 10 base MIs + $\sim$100 scene MIs \\
Level sequences      & Content/Sequences/CV\_Experiments/ & $\sim$100 LevelSequences \\
Pedestrian meshes    & Content/Scanned3DPeoplePack/    & $\sim$620\,MB \\
Weather FX           & Content/Pipeline/FX/            & 3 rain + 3 fog assets \\
Weather blueprints   & Content/Carla/Blueprints/       & 2 (\texttt{BP\_CarlaWeather}, \texttt{BP\_GeneralSceneSettings}) \\
FAB store props      & Content/Fab/                    & misc environmental props \\
\bottomrule
\end{tabular}
\end{table*}

\paraheader{HuggingFace release. } \url{huggingface.co/datasets/allude-occluded/real-carla-assets-anon}. The release excludes the auto-regenerable \texttt{Saved/}, \texttt{Intermediate/}, \texttt{DerivedDataCache/}, directories and only includes relevant asset content (\texttt{Content/},\texttt{Config/}, \texttt{CarlaAssets.uproject} )

\paraheader{Migration to Unreal 5.7. } The most recent version of CARLA's driving content, including town levels, static-mesh, material, and the texture assets, was authored against CARLA's bundled Unreal Engine Fork (Unreal Version 5.5). We re-imported this content into a dedicated 5.7 project using Unreal's forward-compatible content migration, which transfers each asset together with its dependencies while preserving the original paths to avoid breaking any references. All broken CARLA blueprints that failed the content migration were removed from the resulting scene. After the migration, we cleared the \texttt{DerivedDataCache} so that Unreal re-derives shaders and platform data instead of serving stale artifacts from the older engine.

\section{Evaluation}
\label{supp:evaluation}

\subsection{Literature Attacks}
\label{supp:literature-attacks}

Published physical-camouflage attacks couple their texture parameterization, differentiable
rendering bridge, and detector loss into method-specific pipelines that do not transfer across
frameworks. Re-implementing each end-to-end would change the attack, not reproduce it.

We therefore take the published adversarial textures and deploy them, unmodified, on a common
target object (the Audi e-tron) rendered through our framework across a fixed configuration set.
This isolates the rendered, deployed appearance (What the object detector sees) from the specific attack pipeline. \autoref{tab:attacks-evaluated} describes the six evaluated textures: two controls (Benign, Random), three published camouflage attacks (CAMOU, FCA, RAUCA), and our \tool{} texture.

\begin{table*}[h]
\centering
\small
\caption{Descriptions for all 6 evaluated textures: Benign, random, CAMOU, FCA, RAUCA, and \tool}
\label{tab:attacks-evaluated}
\begin{tabular}{p{3cm} p{8cm}}
\toprule
\textbf{Attack} & Description \\
\midrule
Benign & Stock vehicle paint (control). \\
Random & A random-noise texture (control; bounds "any non-trivial texture"). \\
CAMOU & \cite{zhang2018camou} — Learned camouflage; originally optimized against \textbf{Mask R-CNN} in a black-box setting through a learned clone (surrogate) network. Earliest of the three; weakest even against its own detector. \\
FCA & \cite{FCA_2022} (Full-coverage Camouflage Attack) — full-vehicle coverage; originally optimized \textbf{white-box against YOLOv3}. \\
RAUCA & \cite{zhou2024rauca} (Robust And Universal Camouflage Attack) — neural-renderer-based camouflage; originally optimized \textbf{white-box against YOLOv3}; the strongest published baseline. \\
\tool{} & (ours) — A single fixed texture optimized \textbf{natively against DETR} with a targeted (suitcase-class) misclassifications objective at $\epsilon = 256$, $2048^2$ resolution, optimized at the static-elevated / clear-noon viewpoint and applied \textbf{unchanged} to every condition. \\
\bottomrule
\end{tabular}
\end{table*}

\subsection{UV Unification of Literature Attacks}
\label{supp:uv-reunification}

The published camouflage textures are baked for the UV unwrap of its original vehicle mesh.
Applying it to a different mesh whose UV layout differs will scramble the pattern, causing the rendered vehicle to be unfaithful to the original implementation.
To avoid these inconsistencies, we utilize the \texttt{pytorch3d\_Etron} unwrap for the published textures and the \texttt{SM\_EtronParked} texture for \tool{}'s adversarial texture \autoref{tab:etron_texture}.

\begin{table}[h]
\centering
\tiny
\caption{UV layout of the CARLA \texttt{SM\_EtronParked} mesh versus the \texttt{pytorch3d\_Etron} published-texture convention. The two meshes are topologically identical (same face count); they differ only in UV unwrap, which is why the published textures must be re-unified onto the CARLA layout (\autoref{supp:uv-reunification}).}
\label{tab:etron_texture}
\begin{tabular}{lll}
\toprule
 & \textbf{CARLA} 
 \texttt{SM\_EtronParked} 
 & \texttt{pytorch3d\_Etron} \\
\midrule
U range  & $u \in [0,2]$ 
(wrapped / tiled) 
& single $u \in [0,1]$ \\
\addlinespace
Symmetry & left/right mirrored 
(symmetric) 
& asymmetric 
(front $\neq$ rear, L $\neq$ R) 
\\
\addlinespace
Vertices & $20{,}747$ & $13{,}487$ \\
\addlinespace
Topology & $23{,}145$ faces 
(identical) 
& $23{,}145$ faces 
(identical) 
\\
\bottomrule
\end{tabular}
\end{table}

The two meshes share the topology, but differ in their unwrap approach. Applying a published asymmetric texture as symmetric would mirror the adversarial pattern, creating an unfaithful deployment of the published work.
We re-unify the CARLA e-tron body UV to the \texttt{pytorch3d\_Etron} layout so that the published textures are able to be applied exactly as intended to the vehicle.
While the UV layouts of the attacks differ between published attacks and the \tool{} texture, both geometries are identical, allowing direct comparison between attacks.

\paraheader{Method. } We performed the re-UV in Unreal Engine 5.7 through \texttt{GeometryScript}, only editing the \texttt{UV0} and leaving the material slots and scale untouched.
The process is as follows: (1) create a editable dynamic mesh through copying the original CARLA mesh (\texttt{copy\_mesh\_from\_static\_mesh(SM\_EtronParked)}).
(2) Transfer the published work unwrap (\texttt{CARLA\_Etron\_pytorch3dUV.obj}) per triangle through \texttt{set\_mesh\_triangle\_u\_vs} (by value), with a V-flip (\texttt{v\_ue = 1 - v\_obj}) to match the UE texture convention.
(3) Utilize the \texttt{copy\_mesh\_to\_static\_mesh(..., replace\_materials=False)} to write the UV0 back.

\paraheader{Verification. }
After the swap, the body material slot (slot 2) and all other 6 slots are intact, the bounding scale ($485.6 \times 203.3 \times 164.9$\,cm) is identical, and the body UV is now a single asymmetric tile. 
A red-channel / single-texture rendering test produced full asymmetric coverage matching the consumer's reference render, confirming that the unwrap is correct end-to-end.

\subsection{LHS Experiment}
\label{supp:lhs-details}

We model per-cell attack success as a binary outcome (adv-mAP@50 less than 0.5) over the $N=86$ scored cells, using using logistic regression on five categorical factors: optimizer (4 levels), scene–object pair (10 levels), trajectory (5 levels), weather (9 levels), and detector (3 levels).
With reference-level (dummy) encoding this yields 26 predictor columns. Because the predictor count is large relative to N, we apply L2 regularization. For each factor, ``Variance explained'' is the drop in McFadden psuedo-$R^2$ when that factor's columns are removed and the model is refit.
Per factor significance is assess by a likelihood-ratio test between the full model and the model with that factor removed, referenced to a chi-squared distribution with degrees of freedom equal to $d-1$

\subsubsection{Scene Details}
\label{supp:scene-details}
\begin{figure*}[t]
\centering
\includegraphics[width=\linewidth]{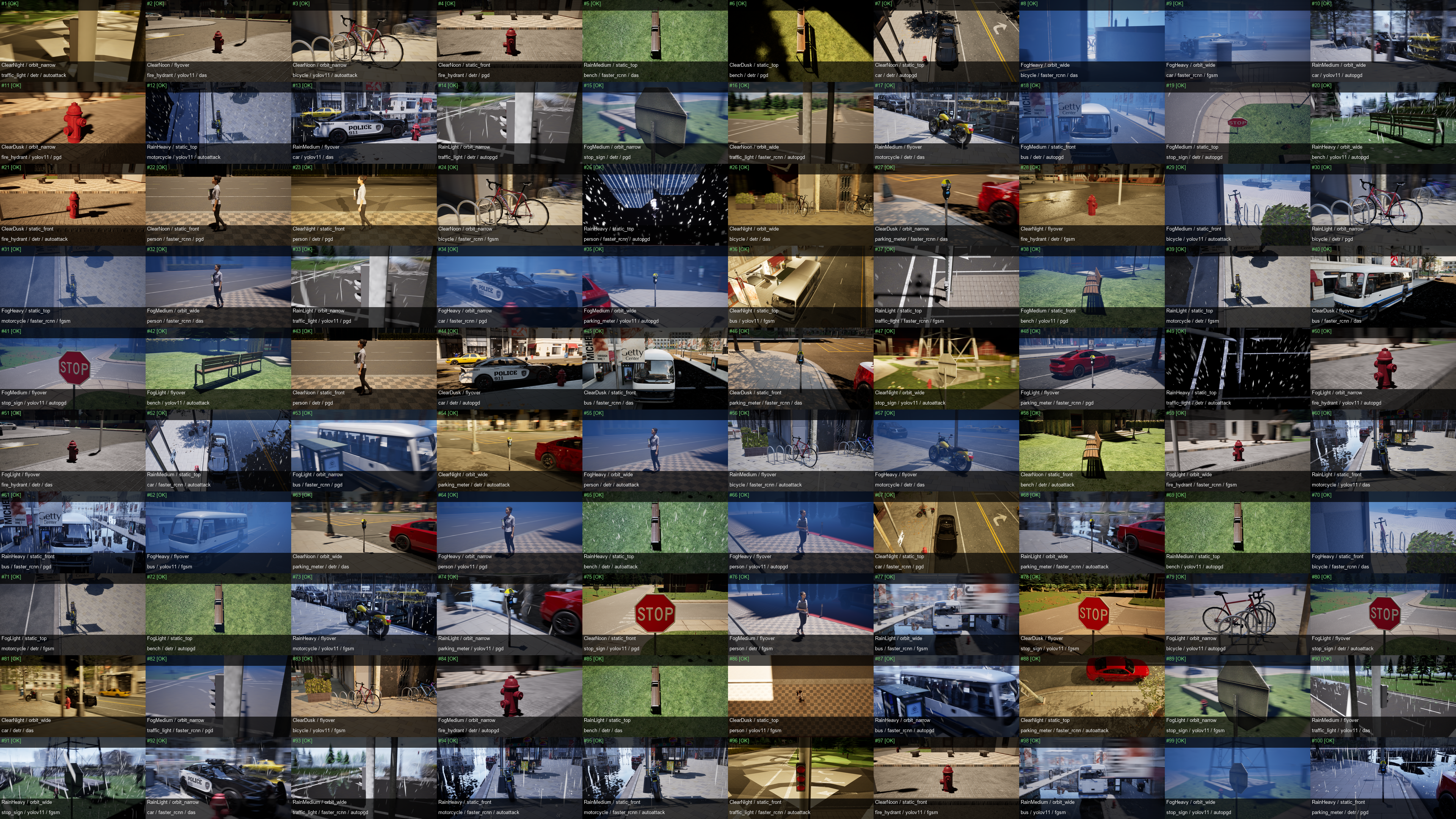}
\caption{Rendered output of all 100 LHS-sampled cells (10 objects $\times$ 10 sampled configurations each), spanning the crossed trajectory and weather conditions.}
\label{fig:lhs-thumbnails}
\end{figure*}

\tool has 10 object, town, and setting context pairs sampled by the LHS. Each scene is crossed with all 5 trajectories $\times$ 9 weather conditions $=$ 45 cells; the balanced LHS samples 10 per object for a total of 100 cells. Refer to \autoref{tab:lhs-scenes} for further details. \autoref{fig:lhs-thumbnails} shows the rendered output of all 100 sampled cells.

\begin{table}[h]
\centering
\small
\caption{The 10 (object, town, setting context) pairs sampled by the LHS. Each scene is crossed with all 5 trajectories $\times$ 9 weather conditions $=$ 45 cells; the balanced LHS samples 10 per object (100 cells total).}
\label{tab:lhs-scenes}
\begin{tabular}{clp{0.30\linewidth}}
\toprule
Object & Town & Setting context \\
\midrule
bench         & 4 & rural park bench \\
bicycle       & 10 & urban curbside bicycle \\
bus           &  10 & parked bus, wide street \\
car           &  10 & parked sedan, downtown stall \\
fire\_hydrant & 2 & small-town sidewalk hydrant \\
motorcycle    &  10 & downtown sidewalk motorbike \\
parking\_meter & 10 & city corner parking meter \\
pedestrian    &  5 & suburban sidewalk person \\
stop\_sign    & 7 & rural intersection sign \\
traffic\_light & 6 & urban-grid intersection \\
\bottomrule
\end{tabular}
\end{table}

\subsubsection{Object Details}
\label{supp:object-details}
Table \autoref{tab:lhs-objects} gives details about the per-object mesh, Material Instance (MI), perturbed material parameter, and UV mask (if any).
\begin{table*}[t]
\centering
\small
\caption{Per-object mesh, Material Instance (MI), and perturbed material parameter.}
\label{tab:lhs-objects}
\begin{tabular}{lclll}
\toprule
Object & COCO & Mesh (\texttt{SM\_*}) & MI
& Param \\
\midrule
bench           & 15 & \texttt{Bench01}                    & \texttt{bench}          & \texttt{BaseColor}         \\
bicycle         &  2 & \texttt{RoadBike}                   & \texttt{bicycle}        & \texttt{AdversarialAlbedo} \\
bus             &  6 & \texttt{MitsubishiFusoRosa\_Parked} & \texttt{bus}            & \texttt{AdversarialAlbedo} \\
car             &  3 & \texttt{DodgeChargerCop\_Parked}    & \texttt{car}            & \texttt{AdversarialAlbedo} \\
fire\_hydrant   & 11 & \texttt{FireHdrant}                 & \texttt{fire\_hydrant}  & \texttt{AdversarialAlbedo} \\
motorcycle      &  4 & \texttt{Harley}                     & \texttt{motorcycle}     & \texttt{Diffuse}           \\
parking\_meter  & 14 & \texttt{Parkingmeter}               & \texttt{parking\_meter} & \texttt{Diffuse}           \\
pedestrian      &  1 & \texttt{rp\_mei\_posed\_001}        & \texttt{pedestrian}     & \texttt{Albedo}            \\
stop\_sign      & 13 & \texttt{Stop01}                     & \texttt{stop\_sign}     & \texttt{SpeedSign\_d}      \\
traffic\_light  & 10 & \texttt{SM\_TrafficeLight\_Merged}  & \texttt{traffic\_light} & \texttt{AdversarialAlbedo} \\
\bottomrule
\end{tabular}
\end{table*}

\section{Dataset Licensing, Impact Statement}
\label{supp:Documentation-impact}

\subsection{Licensing \& Asset Distribution}
\label{supp:documentation}

\tool is open-source under the BSD-3 license. The HuggingFace release (allude-occluded/real-carla-assets-anon) ships only content that is (a) CARLA-derived and CC-BY licensed, or (b) our own original modifications.
Third party-assets governed by non-redistributable licenses, such as Fab-store assets and RenderPeople scanned meshes, have been removed from the release. Users can obtain them separately through their original distributors.

\subsection{Impact Statement}
\label{supp:limitations}

\tool{} enables adversarial attack optimization in a controlled simulation environment. Its primary intended use is defensive, where researchers working on safety-critical-applications can quantify attack surfaces on vision models before deployment.
 
\end{document}